# Predicting 30-Day Hospital Readmission in Medicare Patients
## Insights from an LSTM Deep Learning Model

*Xintao Li
Georgia Institute of Technology
Atlanta, GA, USA
*xli3204@gatech.edu

Sibei Liu
University of Miami
Miami, FL, USA
sxl1086@miami.edu

Dezhi Yu
University of California
Berkeley, CA, USA
dezhi.yu@berkeley.edu

Yang Zhang
MET, Boston University
Boston, MA, USA
ycheung@bu.edu

Xiaoyu Liu
University of Pennsylvania
Philadelphia, PA, USA
xliu833@seas.upenn.edu

*Abstract—* **Readmissions among Medicare beneficiaries are a major problem for the US healthcare system from a perspective of both healthcare operations and patient caregiving outcomes. Our study analyzes Medicare hospital readmissions using LSTM networks with feature engineering to assess feature contributions. We selected variables from admission-level data, inpatient medical history and patient demography. The LSTM model is designed to capture temporal dynamics from admission-level and patient-level data. On a case study on the MIMIC dataset, the LSTM model outperformed the logistic regression baseline, accurately leveraging temporal features to predict readmission. The major features were the Charlson Comorbidity Index, hospital length of stay, the hospital admissions over the past 6 months, while demographic variables were less impactful. This work suggests that LSTM networks offers a more promising approach to improve Medicare patient readmission prediction. It captures temporal interactions in patient databases, enhancing current prediction models for healthcare providers. Adoption of predictive models into clinical practice may be more effective in identifying Medicare patients to provide early and targeted interventions to improve patient outcomes**.

I. INTRODUCTION

Medicare is a federal health insurance program in the United States mainly for people aged 65 or older. Medicare beneficiary readmissions is one of the major challenges in US healthcare system from a perspective of both healthcare operations and patient caregiving outcomes. The Hospital Readmissions Reduction Program (HRRP) created by the US federal Centers for Medicare & Medicaid Services (CMS) serves as a major initiative of penalizing excess 30-day readmissions to the hospitals. The prediction and prevention of senior readmission are now becoming major goals of improving patient outcomes and reducing spending on medical care. The features related to senior readmission provides valuable information of the insight to healthcare provider to optimize the medical decision on the prevention.

There have been various studies focusing on the prediction of hospital readmission, which target specific subpopulations to enhance predictive accuracy. For example, a study on diabetic patients[1] utilized a deep learning model combining wavelet transform and deep forest techniques, achieving an AUROC of 0.726. Similarly, research involving heart failure patients employed random forests with administrative claims data, resulting in an AUC above 0.800. Even though studies with potentially higher predictive accuracy have been published, the LACE index, Length of stay (L), Acuity of admission (A), Comorbidities (C), and recent Emergency department use (E)[2], remains the best-known model for predicting readmissions for general patient populations. Validated through a large-scale study of 1,000,000 Ontarians discharged from hospitals, the LACE index incorporates variables such as length of stay, acuity of admission, comorbidity, and emergency department use, demonstrating solid predictive performance with a C statistic of 0.684.

Significant limitations remain in readmission prediction, as variables like age and gender do not consistently improve model performance, despite the use of scores like the Charlson Comorbidity Index[3]. Variables related to interventions and post-discharge care are often excluded due to data constraints. Additionally, traditional models fail to use sequential EHR data, whereas deep learning models[4], [5], [6], while powerful, can lack the interpretability needed for clinical decision-making[7], [8].

This study aims to develop a deep learning model to address the challenge of predicting hospital readmissions within 30 days after discharge targeting senior Medicare patients. Our approach introduces innovative strategies in data extraction and variable selection, coupled with a recurrent neural network (RNN) architecture with long short-term memory (LSTM) layers that effectively leverages time-series data. To prevent overfitting, we deprioritize patient demographic variables, as they offer minimal contribution to model performance[9]. Instead, our model integrates admission-level data, such as surgeries, consultations, and medications, with patient-level data before and after discharge, including ICD codes and Emergency Department admission history. The LSTM model's performance is compared with a logistic regression model incorporating the LACE index risk score, allowing us to compare its effectiveness in predicting readmissions among Medicare patients.

## II. Model Development

### A. Baseline Model: LACE Index

To get a robust baseline model for senior readmission prediction, we developed a multivariate logistic regression model with LACE index components[2]. The LACE index components consist of Length of stay (L), Acuity of the admission (A), Comorbidities (C), and Emergency department visits (E). Logistic regression is a statistical method that models the binary outcomes of one or more independent variables. The logistic regression model is of the form as:

$$\text{logit}(p) = \beta_0 + \beta_1 X_1 + \beta_2 X_2 + \cdots + \beta_n \quad (1)$$

In this equation, $p$ represents the probability of readmission, $\beta_0$ is the intercept, and $\beta_i$ are the coefficients corresponding to the predictor variables $X_i$. To enhance the interpretability of the model, we adopted the approach suggested by Sullivan et al.[10] to transform the logistic regression model into a risk index. This step involves converting a model's regression coefficients into point scores by taking each coefficient and dividing by the smallest absolute value of any coefficient in the model, and then rounding the result off to the nearest whole number. The overall risk score for an individual patient is the sum of those points.

The logistic regression model was trained and validated using a 70-30 train-test split, allowing us to provide a benchmark for more complex models that could be trained by using deep learning techniques.

TABLE I. Components and Scoring of the LACE Index for 30-Day Readmission Risk

| LACE Component | Criteria | Score |
|---|---|---|
| Length Of Stay (L) | < 1 | 0 |
|  | 1 | 1 |
|  | 2 | 2 |
|  | 3 | 3 |
|  | 4-6 | 4 |
|  | 7-13 | 6 |
|  | >14 | 7 |
| Acuity of Admission (A) | Yes | 3 |
| Charlson Comorbidity Index (C) | 0 | 0 |
|  | 1 | 1 |
|  | 2 | 2 |
|  | 3 | 3 |
|  | >=4 | 5 |
|  | 0 | 0 |
| Visits To Emergency Department (E) In Previous 6 Months | 1 | 1 |
|  | 2 | 2 |
|  | 3 | 3 |
|  | >=4 | 4 |

### B. Long short-term memory (LSTM) model

We utilized Long Short-Term Memory (LSTM) networks to build our predictive model on time series data. The clinical data from patients are usually available at irregular intervals and require missing-value imputation[11]. The LSTM networks were particularly a good fit for temporal dependencies and patterns[12]. LSTM networks is a specialized form of Recurrent Neural Networks (RNNs) that designed to manage long-term dependencies in sequential data[13], [14]. The unique architecture composed of a series of gates: forget, input, and output gates, which regulate the information flow and ensure that relevant details are retained over extended periods[11]. The LSTM is of the form as

$$f_t = \sigma(W_f \cdot [h_{t-1}, x_t] + b_f) \quad (2)$$

$$i_t = \sigma(W_i \cdot [h_{t-1}, x_t] + b_i) \quad (3)$$

$$\widetilde{C}_t = \tanh(W_C \cdot [h_{t-1}, x_t] + b_C) \quad (4)$$

$$C_t = f_t * C_{t-1} + i_t * \widetilde{C}_t \quad (5)$$

$$o_t = \sigma(W_o \cdot [h_{t-1}, x_t] + b_O) \quad (6)$$

Here, $f_t$ is the forget gate, $i_t$ is the input gate, $\widetilde{C}_t$ is the candidate cell state, $C_t$ is the cell state, $o_t$ is the output gate, and $h_t$ is the hidden state. The sigmoid function ($\sigma$) and hyperbolic tangent function (tanh) introduce non-linearity to the model.

Our LSTM model architecture consists of a bidirectional LSTM layer and an additional LSTM layer, allowing for the model to adapt to patterns in both forward and backward directions in the time-series data. Because temporal dynamics can be complex in many situations, dual-layer LSTMs are better at accommodating nonlinear interactions between how future and past events influence each other. The decimal forecast was assigned to the single output neuron in a dense decision layer, activated by a sigmoid function, which is used for binary classification.

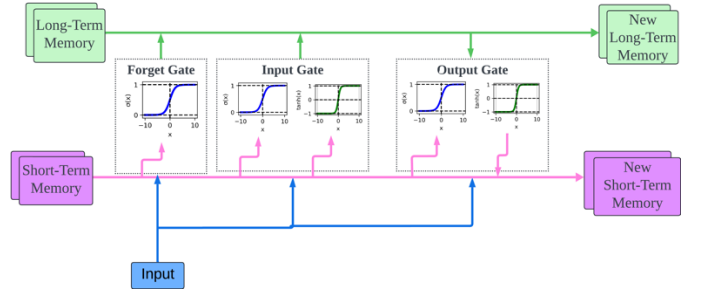

Figure 1. Long Short-Term Memory Neural Networks

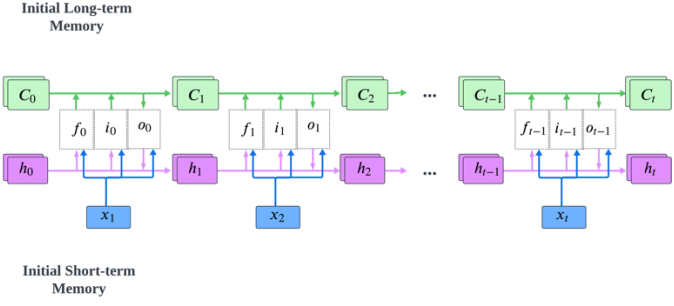

Figure 2.  Long Short-Term Memory Flow Chart

Fig. 1 and Fig. 2 illustrate how LSTM does work on top of RNN. Fig. 1 is a sequential model of LSTM that works on each time step until it reaches the end of the data. Every step of the process has input that can influence the short-term and long-term memory, while forget, input, and output gates regulate the information flow [13]. The interplay between these gates ensures that the LSTM can selectively remember and utilize relevant information over long sequences. This is crucial for tasks of time-series data or sequential information processing of any form. Fig. 2 exhibits how in each time step, it depends on the others to capture some complicated temporal pattern in the sequence data. By remembering and updating the long-term memory state of the dependency through time flow, the LSTM can avoid vanishing gradient, which occurs in a traditional RNN. Thus, LSTM provides a robust way to manage the long-term dependency in the input data.

To optimize the model, we used the Adam optimizer with a learning rate of 0.001, selected through hyperparameter tuning. The Adam optimizer works well on datasets with sparse gradients and noisy data like our clinical dataset [15]. We used binary cross-entropy as the loss function in equation (7) to measure the discrepancy between the predicted and actual values:

$$L(y, \hat{y}) = -\frac{1}{N}\sum_{i=1}^{N}[y_i \log(\hat{y}_i) + (1 - y_i)\log(1-\hat{y}_i)] \quad (7)$$

where $y$ is the true label, $\hat{y}$ is the predicted probability, and $N$ is the number of samples. Additionally, we used the Area Under the Curve (AUC) metric to evaluate the model's performance. We applied layer-wise dropout (rate of 0.4) to prevent overfitting [16] and used Early Stopping to halt training when validation loss stopped improving. The LSTM model was trained with a 70-30 train-test split.

### III. EVALUATION AND RESULTS

#### A. Evaluation

The dataset for this study was from the MIMIC-III database[17], which comprises data for 53,423 unique hospital admissions of adult patients between 2001 and 2012. We cleaned and transformed the data into 21002 Medicare patients[18]. We set the readmission window to 30-days and utilized several key metrics for the evaluation of model performance. The Area Under the Receiver Operating Characteristic Curve (AUC-ROC) serves as the primary measure of predictive accuracy, evaluating a model's ability to distinguish between positive (readmitted) and negative (not readmitted) classes. A higher AUC indicates superior model performance at correctly classifying patients. Additionally, we focused on precision and recall within the "highest risk decile" of patients. This involved ranking patients according to their predicted risk scores and analyzing the top 10%, those deemed most at risk[19]. Precision here is a measure that reflects the proportion of patients identified as high-risk who were readmitted. Recall, conversely, measures the proportion of actual readmissions captured within this high-risk group, indicating the model's effectiveness in identifying the majority of true readmissions.

$$\text{Precision} = \frac{\text{True Positives}}{\text{True Positives+False Positives}} \quad (8)$$

$$\text{Recall} = \frac{\text{True Positives}}{\text{True Positives+False Negatives}} \quad (9)$$

High precision ensures that most patients flagged as high-risk are indeed readmitted, reducing unnecessary interventions. High recall, on the other hand, ensures that the model successfully identifies a substantial proportion of all actual readmissions, thereby minimizing missed cases[16]. To ensure a robust evaluation, these metrics were averaged over 20 different data splits[20], with random assignments to reduce the potential for overfitting.

#### B. Results

**Baseline Model:** The baseline LACE index model using logistic regression achieved an AUC of 0.608 (95% CI: 0.602 - 0.614), lower than the reported AUC of 0.85 in a general population[21] . For high-risk readmissions, the model had a precision of 0.168 (95% CI: 0.161 - 0.175) and recall of 0.261 (95% CI: 0.252 - 0.270), meaning 16.8% of predicted high-risk patients were readmitted, while only 26.1% of actual readmissions were identified. While doing features importance analysis with Odds Ratio (OR), key predictors were length of hospital stay (OR 1.015), increased Charlson comorbidity score (OR 1.127), and prior readmissions within six months (OR 1.199). The odds of readmission among females were 0.802 compared to males. The results show that reducing readmissions will likely require interventions focusing on these predictors.

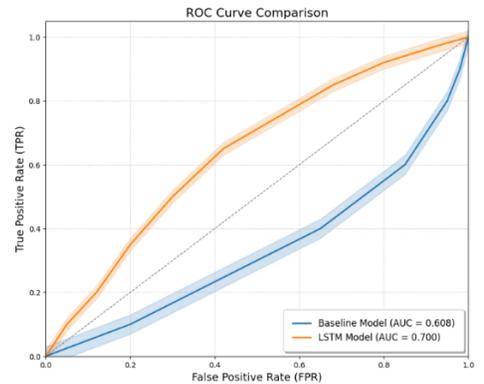

Figure 3.  AUC Comparison of Logistic Regression and LSTM

**LSTM Model**: The LSTM model demonstrated notable performance improvements over the baseline logistic regression model as shown in Fig. 3, achieving an Average AUC of 0.700 (95% CI: 0.693 - 0.706), compared to the logistic regression's Average AUC of 0.608 (95% CI: 0.602 - 0.614). Using precision and recall evaluating the model performance in terms of top 10% high-risk admission patients, LSTM model achieved precision of 0.355 (95% CI: 0.319 – 0.391) and recall of 0.418 (95% CI: 0.389 – 0.448), a significant improvement from the baseline model.

**Features Analysis:** We utilized permutation importance measured in AUC to further understand the contributions of different features in predicting hospital readmissions within 30 days as shown in Fig. 4. Permutation importance measures the decrease in model performance when a single feature's values are randomly shuffled, effectively breaking the relationship between that feature and the target variable. Our analysis reveals that the Charlson Comorbidity Index (CCI) score is the most significant predictor, with an importance value of 0.1424. The second most important feature is the length of hospital stay, highlighting the critical role of prolonged hospitalizations in predicting readmissions. Interestingly, some features exhibit negative importance values, suggesting they may have a negligible or potentially misleading impact on the model's predictions. For instance, age has an importance value of -0.0040, and admissions for surgery have an importance value of -0.0007. These negative values indicate that these features, when permuted, might slightly improve the model's performance, thus questioning their relevance.

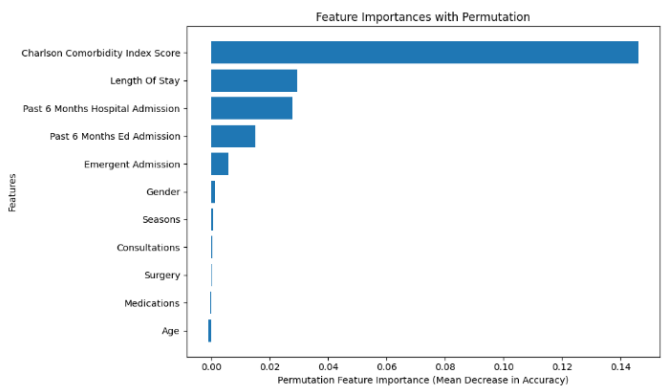

Figure 4.  Permutation Analysis

**Model Adjustment:** Recognizing this, we undertook feature selection and engineering to refine our model further. The results are listed in Table 3. In our feature selection process, removing the Age feature alone resulted in an Average AUC of 0.706 (95% CI: 0.701 - 0.711), maintaining precision but slightly reducing recall. Removing the Surgery feature led to a decrease in performance with an Average AUC of 0.695 (95% CI: 0.690 - 0.700). Combining the removal of both features yielded an Average AUC of 0.699 (95% CI: 0.694 - 0.705), indicating the detrimental impact of excluding these variables together. Next, we explored feature engineering by categorizing patient ages[22],[23] into ranges: 65-74, 74-85, and 85+. This categorization improved the model's Average AUC to 0.703 (95% CI: 0.697 - 0.710), with precision remaining stable and a slight increase in recall. Combining this age categorization with the removal of the Surgery feature resulted in a diminished performance, with an Average AUC of 0.694 (95% CI: 0.690 - 0.699). Ultimately, our best-performing LSTM model excluded the Age feature but retained the Surgery variable in its original form, achieving an optimal balance between predictive accuracy and recall. These findings highlight the nuanced impacts of feature selection and engineering on model performance, emphasizing the importance of retaining critical variables for robust predictions.

## IV. Discussion

### A. In-Depth Analysis of Variable Contributions to Readmission Risk: Insights from SHAP

Using SHapley Additive exPlanations (SHAP) as shown in Fig. 5, we found that the Charlson Comorbidity Index (CCI) is the most significant predictor of 30-day readmissions, followed by hospital length of stay, which reflects illness severity. Surprisingly, more medications and consultations during initial hospitalization were linked to a lower readmission risk, suggesting better care reduces returns to the hospital. We also observed higher readmission rates among female patients, likely due to physiological and care-related differences[24]. Prior emergency visits were moderate predictors of readmission, while surgeries and seasonal variations had little effect.

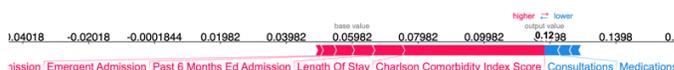

Figure 5.  SHAP Force Plots

### B. Clinical Validation of Model Predictors

We reference studies that affirm key LSTM model predictors to enhance clinical relevance. The Charlson Comorbidity Index (CCI) stands out as a crucial predictor, supported by Dongmei's research on heart failure patients, which shows higher CCI scores correlate with increased readmission risks[25]. Canepa's study also links elevated CCI with higher readmissions in elderly patients[26]. Jean-Sebastien's analysis reinforces the significance of length of stay (LOS) on readmission risk[27], [28]. Prior hospitalizations, particularly via emergency departments, are strong predictors of rehospitalization within 30 days. These factors likely reflect the overall illness burden and social conditions[2]. Conversely, the model's finding that more discharge medications reduce readmission risk is supported by Picker's study[29], while consultations at discharge also lower readmission rates. Gender-specific predictions align with research on physiological differences that increase complication rates in women[30], such as smaller body size and arteries. Removing age as a variable improved model performance, consistent with research showing age's influence stabilizes after 65[31]. These findings confirm the LSTM model's predictive accuracy and clinical value, merging machine learning with clinical data to improve care and reduce readmissions.

## V. CONCLUSIONS

This study underscores the significance of predictive modeling in addressing the challenge of 30-day hospital readmissions among senior Medicare patients. By leveraging a Long Short-Term Memory (LSTM) model, we have demonstrated that incorporating time-series data and focusing on admission-level variables enhances the prediction of readmission risk, surpassing traditional logistic regression models that rely on the LACE index.

The findings highlight that the Charlson Comorbidity Index (CCI) and length of hospital stay are the most influential factors in predicting readmissions. Our LSTM model's ability to capture temporal dependencies allows it to provide a more nuanced understanding of these predictors compared to baseline models. Moreover, the model's precision and recall metrics in identifying high-risk patients indicate its potential utility in clinical settings for targeted interventions.

Overall, this research contributes to the ongoing efforts to reduce hospital readmissions by providing a robust and clinically relevant predictive model. The insights can inform healthcare providers and policymakers in developing strategies to improve patient outcomes and reduce healthcare costs. As the healthcare landscape continues to evolve, the integration of advanced machine learning models like the one presented here will be crucial in driving forward data-driven, patient-centered care.